\pdfoutput=1

\documentclass[11pt]{article}

\usepackage{acl}

\usepackage{times}
\usepackage{latexsym}

\usepackage[T1]{fontenc}

\usepackage[utf8]{inputenc}

\usepackage{inconsolata}

\usepackage{setspace}
\usepackage{hyperref}
\usepackage{bm}
\usepackage{bbding}
\usepackage{color}
\usepackage{enumitem}
\usepackage{amsmath}
\usepackage{booktabs}
\usepackage{graphicx}
\graphicspath{{figure/}}
\DeclareGraphicsExtensions{.pdf,.png}
\usepackage{subcaption}

\title{SemEval-2024 Task 3: Multimodal Emotion Cause Analysis\\ in Conversations}

\author{Fanfan Wang$^{1}$, Heqing Ma$^{1}$, Jianfei Yu$^{1*}$, Rui Xia$^{1}$\thanks{\hspace{0.3em} Corresponding authors.}, Erik Cambria$^{2}$\\ 
  $^{1}$ School of Computer Science and Engineering, \\Nanjing University of Science and Technology, China\\
  $^{2}$ Nanyang Technological University, Singapore \\
  \texttt{\{ffwang, hqma, jfyu, rxia\}@njust.edu.cn}, \texttt{cambria@ntu.edu.sg}
}

\begin{document}
\maketitle
\begin{abstract}
The ability to understand emotions is an essential component of human-like artificial intelligence, as emotions greatly influence human cognition, decision making, and social interactions. 
In addition to emotion recognition in conversations,  
the task of identifying the potential causes behind an individual's emotional state in conversations, is of great importance in many application scenarios. 
We organize SemEval-2024 Task 3, named Multimodal Emotion Cause Analysis in Conversations, which aims at extracting all pairs of emotions and their corresponding causes from conversations. Under different modality settings, it consists of two subtasks: Textual Emotion-Cause Pair Extraction in Conversations (TECPE) and Multimodal Emotion-Cause Pair Extraction in Conversations (MECPE).
The shared task has attracted 143 registrations and 216 successful submissions.
In this paper, we introduce the task, dataset and evaluation settings, summarize the systems of the top teams, and discuss the findings of the participants. 
\end{abstract}

\section{Introduction}

Understanding emotions is crucial to achieve human-like artificial intelligence, as emotions are intrinsic to humans and significantly influence our cognition, decision-making, and social interactions. 
Conversation is an important form of human communication and contains a large number of emotions.
Furthermore, given that conversation in its natural form is multimodal, many studies have explored multimodal emotion recognition in conversations (ERC), using language, audio and vision modalities \cite{poria2019emotion,mittal2020m3er,lian2021ctnet,zhao2022m3ed,zheng2023facial}.

However, emotion recognition alone is not sufficient to fully understand the intricacies of human emotions. Emotion cause analysis (ECA), the process of identifying the potential causes behind an individual's emotion state, has broad application scenarios such as human-computer interaction, commerce customer service, empathetic conversational agents, and automatic psychotherapy.
For example, conversational agents equipped with emotion cause analysis can better understand the user's emotional state, offer empathetic responses, and provide more personalized services.
By identifying the cause of the emotional state of a patient, a psychotherapy system can provide more accurate and customized treatments.
ECA has gained increasing attention both in academic and practical fields \cite{ding2019independent,Xia2019rthn,xia2019emotion,ding2020ecpe,ding2020end,poria2021recognizing,li2022ecpec,an2023global,wang2023generative}. 
However, to our knowledge, there has not been any evaluation competition conducted specifically for emotion cause analysis in conversations.

To promote research in this direction, we organize a shared task in SemEval-2024, named Multimodal Emotion Cause Analysis in Conversations.
Our task consists of two subtasks: Subtask 1 (Textual Emotion-Cause Pair Extraction in Conversations, TECPE) focuses on extracting emotion and textual cause spans solely based on text; Subtask 2 (Multimodal Emotion-Cause Pair Extraction in Conversations, MECPE) involves extracting emotion-cause pairs at the utterance level considering three
modalities.

\begin{figure*}[t]
	\centering
	\includegraphics[width=\textwidth]{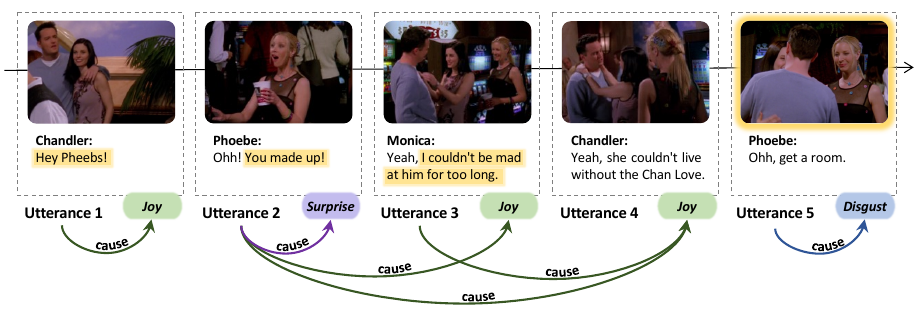}
	\caption{An example of our task and annotated dataset. 
 Each arc points from the cause utterance to the emotion it triggers. The textual cause spans and the visual cause evidence are highlighted in yellow.
 Background: Chandler and his girlfriend Monica walked into the casino (they had a quarrel earlier but made up soon) and then started a conversation with Phoebe.
}
	\label{fig:example}
\end{figure*}

For this shared task, we provide a multimodal emotion cause dataset ECF 2.0 sourced from the sitcom \textit{Friends}. This dataset contains 1,715 conversations and 16,720 utterances, where 12,256 emotion-cause pairs are annotated at the utterance level, covering three modalities (language, audio, and vision). 
Specifically, in our preliminary work \cite{wang2022multimodal}, we have constructed a benchmark dataset, Emotion-Cause-in-Friends (ECF 1.0), which contains 1,374 conversations and 13,619 utterances. 
On this basis, we have furthermore annotated an extended test set as the evaluation data and provided the span-level annotations of emotion causes within the textual modality.

Our task has attracted 143 registrations and a total of 216 successful submissions during the 16-day evaluation phase. 
Participants tended to decompose our task into emotion recognition and cause prediction, proposing numerous well-designed pipeline systems. Moreover, many teams applied advanced Large Language Models (LLMs) for emotion cause analysis and achieved promising results.
After the evaluation, 18 teams finally submitted system description papers.


\section{Task}
\label{sec:task}

We clarify the definitions of emotion and cause before introducing the task and dataset.
\textbf{Emotion} is a psychological state associated with thought, feeling, and behavioral response \cite{ekman1994nature}. In computer science, emotions are often described as discrete emotion categories, such as Ekman’s six basic emotions, including \textit{Anger}, \textit{Disgust}, \textit{Fear}, \textit{Joy}, \textit{Sadness} and \textit{Surprise} \cite{1971Universals}. 
In conversations, emotions were usually annotated at the utterance level \cite{li2017dailydialog, hsu2018emotionlines,poria2019meld}.
\textbf{Cause} refers to the objective event or subjective argument that triggers the corresponding emotion \cite{lee2010text,russo2011emocause}.

The goal of our shared task, named Multimodal Emotion Cause Analysis in Conversations, is to extract potential pairs of emotions and their corresponding causes from a given conversation. 
Figure \ref{fig:example} illustrates a typical multimodal conversation scenario, which involves multiple emotions and their corresponding causes. 
Under different modality settings, we define the following two subtasks:

\paragraph{Subtask 1: Textual Emotion-Cause Pair Extraction in Conversations (TECPE).}
Extracting all emotion-cause pairs from the given conversation solely based on text, where each pair contains an emotion utterance along with its emotion category and the textual cause span, e.g., (\textit{U3}\_\textit{Joy}, \textit{U2}\_``\textit{You made up!}'') in Figure \ref{fig:example}.

\paragraph{Subtask 2: Multimodal Emotion-Cause Pair Extraction in Conversations (MECPE).}
It should be noted that sometimes the cause cannot be reflected only in text. 
As shown in Figure \ref{fig:example}, the cause for Phoebe’s \textit{Disgust} in \textit{U5} is that Monica and Chandler were kissing in front of her, which is reflected in the visual modality of \textit{U5}. 
Therefore, we accordingly define this multimodal subtask to extract all emotion-cause pairs in consideration of three modalities (language, audio, and vision). In this subtask, the cause is defined at the utterance level, and each pair contains an emotion utterance along with its emotion category and a cause utterance, e.g., (\textit{U5}\_\textit{Disgust}, \textit{U5}).

\section{Dataset}
\label{sec:dataset}

\subsection{Data Source}

Sitcoms come with real-world-inspired inter-human interactions and usually contain more emotions than other TV series or movies. 
Based on the famous American sitcom \textit{Friends}, \citet{poria2019meld} constructed the multimodal conversational dataset MELD by extracting audiovisual clips corresponding to the scripts of the source episodes and annotating each utterance with one of six basic emotions (\textit{Anger}, \textit{Disgust}, \textit{Fear}, \textit{Joy}, \textit{Sadness} and \textit{Surprise}) or \textit{Neutral}. 
MELD has recently become a widely used benchmark for ERC.

In our preliminary work \cite{wang2022multimodal}, we chose MELD as the data source and further annotated the causes given emotion annotations, thereby constructing the ECF 1.0 dataset. 
For this SemEval competition, we release the entire ECF 1.0 dataset as a training set and additionally create a test set as evaluation data, which is also sourced from \textit{Friends}.

\subsection{Data Collection}

To construct the extended test set, we first crawl the subtitle files of all the episodes of \textit{ Friends}, which contains the utterance text and the corresponding timestamps.
The subtitles are then separated by scene (scene descriptions are written in square brackets in the subtitle files), and each scene in every episode is viewed as a conversation. If the length of a conversation exceeds 40 utterances, we further divide it into several conversations of random lengths.
Conversations included in the ECF 1.0 are removed.
Next, we divide the collected conversations into several parts according to their lengths, with each part falling within the length ranges [1, 5], [6, 10], [11, 15], [16, 20], [21, 25], and [26, 35], respectively.
Finally, we randomly sample conversations from each part according to the distribution probability of conversation lengths in ECF 1.0, and a total of 400 conversations are sampled for annotation.

\subsection{Data Annotation}

We employ three graduate students involved in the annotation of the ECF 1.0 dataset to annotate the extended test set. Given a multimodal conversation, they first need to annotate the speaker and emotion category for each utterance, and then further annotate the utterances containing corresponding causes for each non-neutral emotion. If the causes are explicitly expressed in the text, they should also mark the textual cause spans.
After annotation, we determine the emotion categories and cause utterances by majority voting, and take the largest boundary (i.e., the union of the spans) as the gold annotation of the textual cause span. If disagreements arise, another expert is invited for the final decision.


\begin{table}[t]
\centering
\small
\renewcommand{\arraystretch}{1.2}
\resizebox{0.48\textwidth}{!}
{
\begin{tabular}{lcccr}
\hline
\textbf{Dataset} & \textbf{Modality} & \textbf{Scene} & \textbf{ \# Ins} \\
\hline
Emotion-Stimulus  \cite{ghazi2015detecting} & T & -- & 2,414 s \\
ECE Corpus \cite{gui2016event} & T & News & 2,105 d \\
NTCIR-13-ECA  \cite{gao2017overview} & T & Fiction & 2,403 d \\
Weibo-Emotion \cite{cheng2017emotion} & T & Blog & 7,000 p \\
REMAN \cite{kim2018feels} & T & Fiction & 1,720 d \\
GoodNewsEveryone \cite{bostan2020goodnewseveryone} & T & News & 5,000 s \\
RECCON-IE \cite{poria2021recognizing} & T & Conv & 665 u \\
RECCON-DD \cite{poria2021recognizing} & T & Conv & 11,104 u \\
ConvECPE \cite{li2022ecpec} & T,A,V & Conv & 7,433 u \\
\hline
ECF 1.0 \cite{wang2022multimodal}  & T,A,V & Conv & 13,619 u \\ 
\textbf{ECF 2.0} & T,A,V & Conv & 16,720 u \\ 
\hline
\end{tabular}
}
\caption{Comparison of existing ECA datasets. T, A, and V refer to text, audio, and video. Blog and Conv represent microblog and conversation, and s, d, p and u denote sentence, document, post and utterance.}
\label{tab:datasets}
\end{table}

\begin{table}[t]
\centering
\resizebox{0.48\textwidth}{!}
{
\begin{tabular}{lccc}
\hline
    \textbf{Items} & \textbf{ECF 1.0} & \textbf{Extended Test} & \textbf{ECF 2.0} \\ \hline
    Conversations & 1,374 & 341 & 1,715 \\
    Utterances & 13,619 & 3,101 & 16,720 \\
    Emotion (utterances) & 7,690 & 1,821 & 9,511 \\
    \hline
    \multicolumn{4}{c}{\textbf{Subtask 1 (TECPE)}} \\
    \hline
    Emotion (utterances) with causes & 6,761 & 1,626 & 8,387 \\
    Emotion-cause (span) pairs & 9,284 & 2,256 & 11,540 \\
    \hline
    \multicolumn{4}{c}{\textbf{Subtask 2 (MECPE)}} \\
    \hline
    Emotion (utterances) with causes & 7,081 & 1,746 & 8,827 \\
    Emotion-cause (utterance) pairs & 9,794 & 2,462 & 12,256 \\
\hline
\end{tabular}
}
\caption{Statistics of our dataset.}
\label{tab:ECF2}
\end{table}

\paragraph{Annotation Cost. }
The average duration of each conversation in our dataset is 31.6 seconds and it takes about 10 minutes to annotate a conversation. 
Each annotator would be paid CNY 300 when finishing every 50 conversations, which leads to the basic salary of CNY 36 (USD 5.2) per hour, which is higher than the current average salary in Jiangsu Province, China.

\paragraph{Data Post-processing.}
We conduct the following post-processing and cleaning of the data:
\begin{itemize}
\setlength\itemsep{-0.2em}
\item Correct the utterance text that does not match what the speaker said in the video;
\item Correct the timestamps that are not aligned with utterance text;
\item Separate the utterance whose segment of timestamps covers two speakers' utterances and modify their timestamps;
\item Separate the conversation which spans scenes;
\item Discard conversations if there is significant disagreement in annotations and the expert also finds it difficult to determine.
\end{itemize}
After these steps, we store the text data in JSON files separately for each subtask. For Subtask 2, we use the FFmpeg\footnote{\url{https://www.ffmpeg.org}} tool to extract video clips of each utterance from the source episodes based on the start and end timestamps.

\begin{figure}[t]
\centering
\subfloat[ECF 1.0]{\includegraphics[width = 0.45\textwidth]{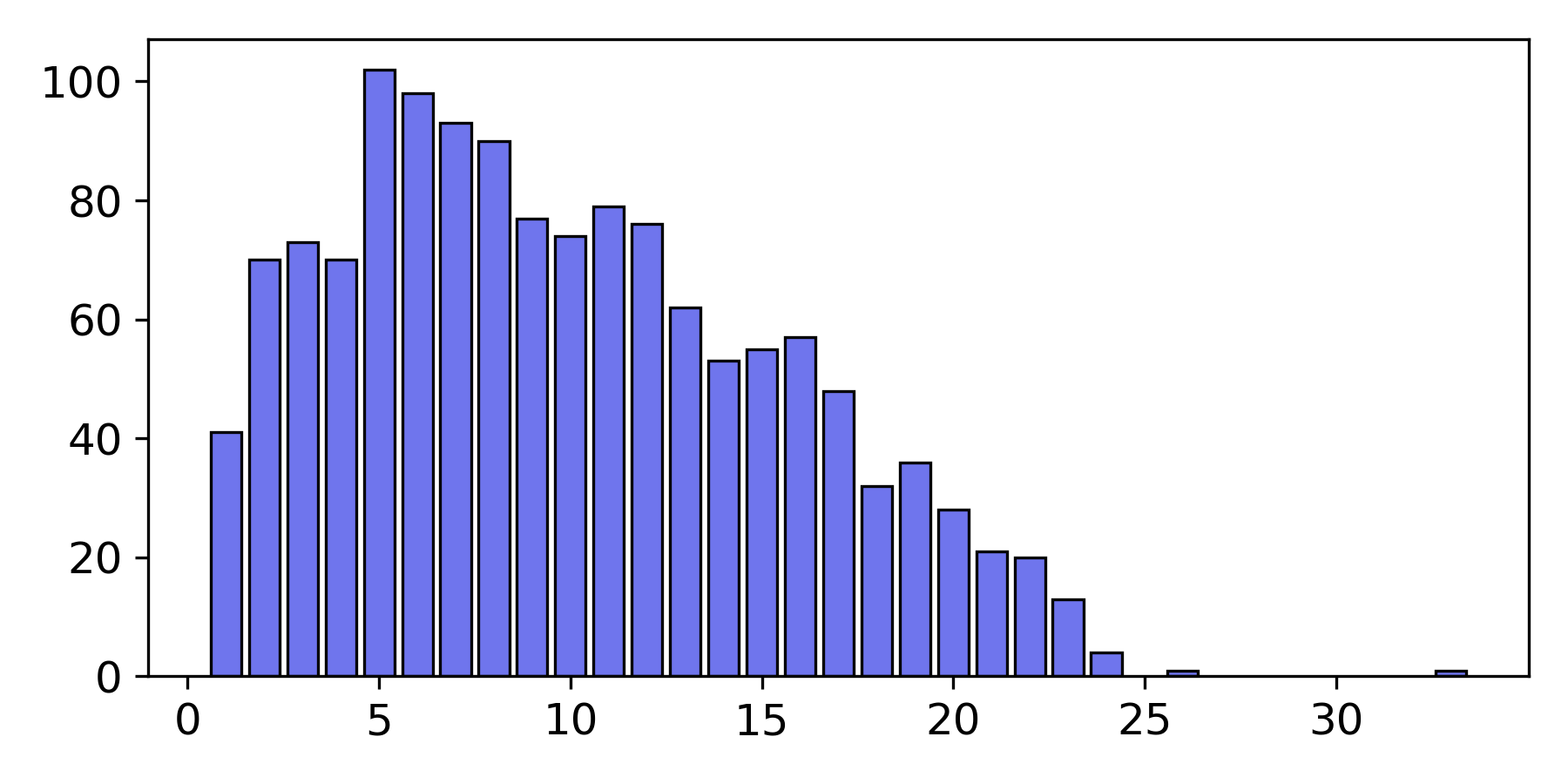}
\label{fig:a}}
\hfil
\subfloat[Extended Test set for SemEval-2024]{\includegraphics[width = 0.45\textwidth]{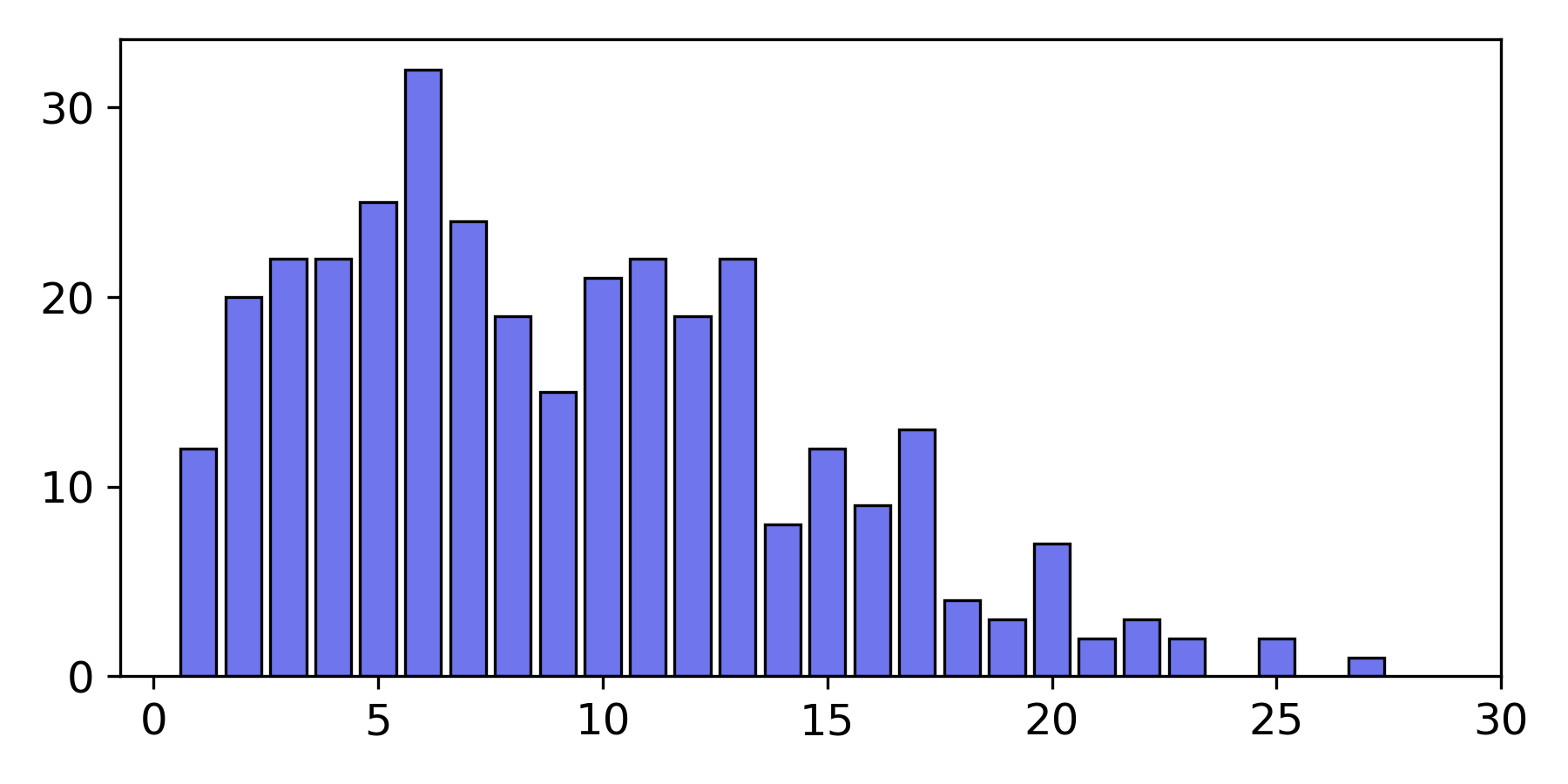}
\label{fig:b}}
\caption{The distribution of conversation lengths. The horizontal axis represents the number of utterances, and the vertical axis represents the number of conversations.}
\label{fig:compare_conv_length}
\end{figure}

\begin{figure}[t]
\centering
\subfloat[ECF 1.0]{\includegraphics[width = 0.45\textwidth]{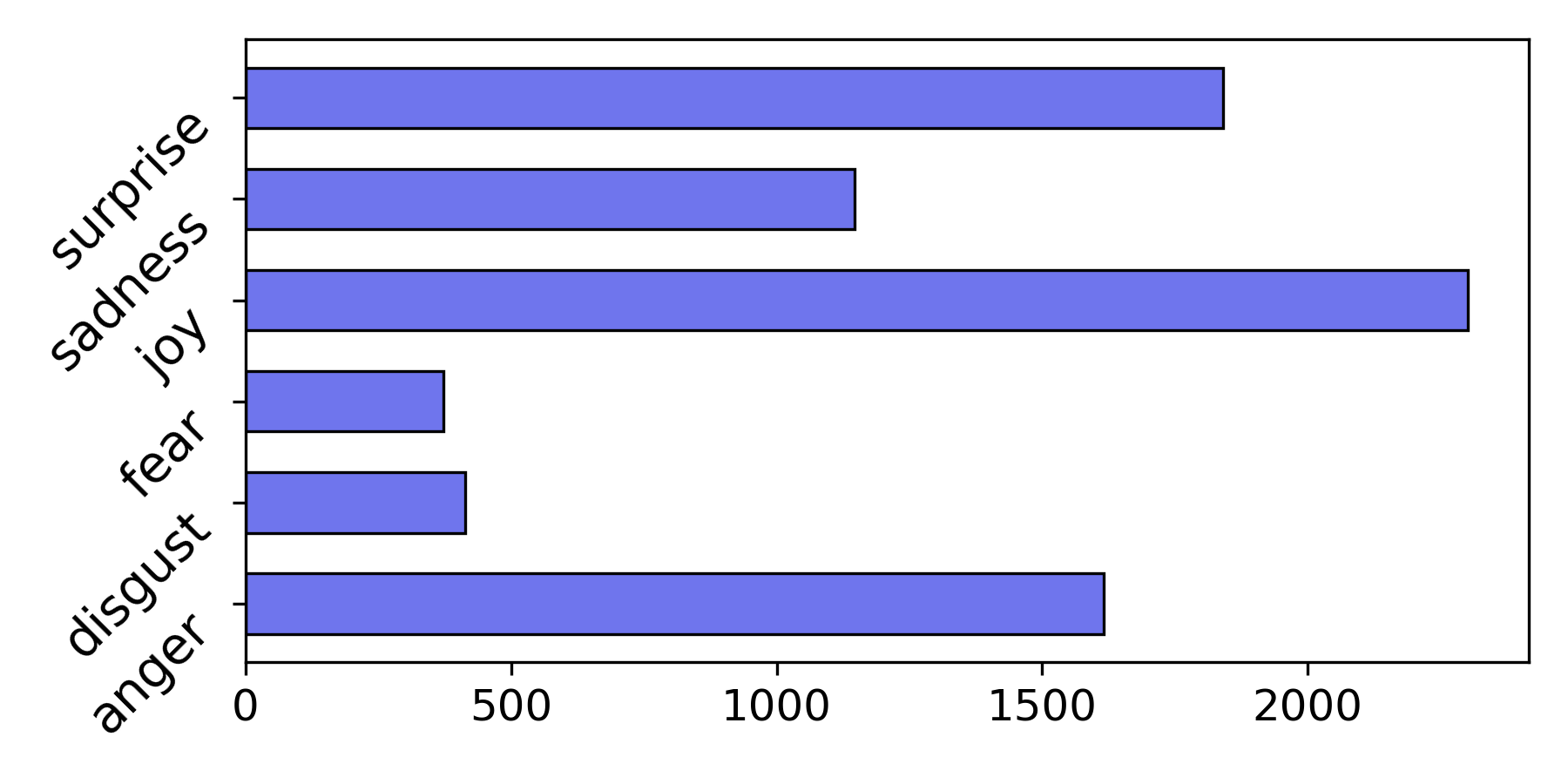}
\label{fig:emo_a}}
\hfil
\subfloat[Extended Test set for SemEval-2024]{\includegraphics[width = 0.45\textwidth]{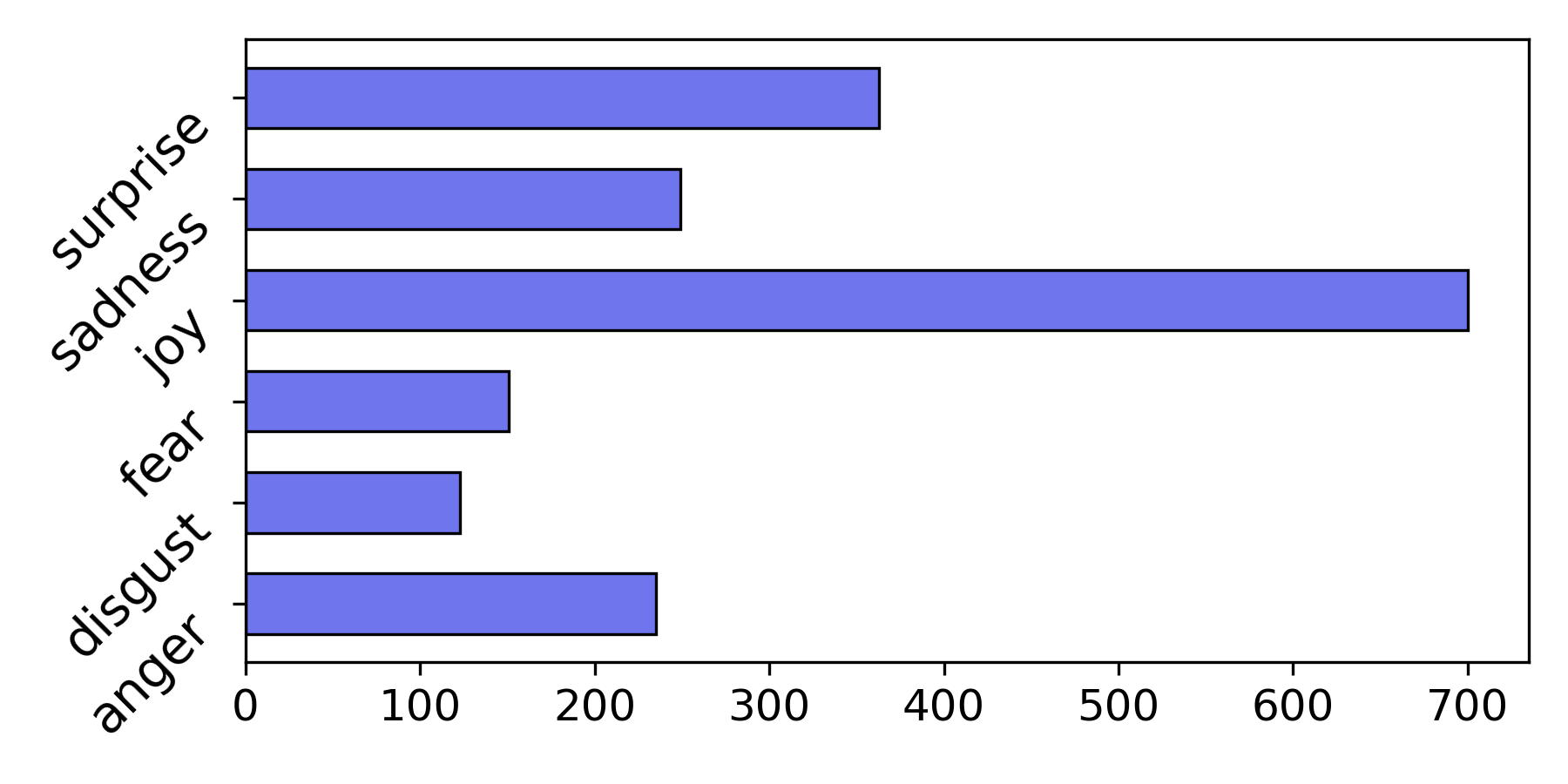}
\label{fig:emo_b}}
\caption{The distribution of emotions. The horizontal axis represents the number of utterances, and the vertical axis represents emotion categories.}
\label{fig:compare_emotion}
\end{figure}

\subsection{Dataset Statistic}
In our preliminary work \cite{wang2022multimodal}, we have already constructed the ECF 1.0 dataset that contains 1,374 conversations and 13,619 utterances.
Furthermore, we have annotated an extended test set specifically for this SemEval evaluation, which together with ECF 1.0 constitutes the \textbf{ECF 2.0} dataset\footnote{Our dataset is available on \href{https://drive.google.com/drive/folders/1TIRBiL8z4ZnoxtuKM8pnjtm2BxB5mS4Y}{Google Drive}.}  that contains 1,715 conversations and 16,720 utterances.

In Table \ref{tab:datasets}, we compare our dataset with the related datasets for ECA, in terms of modality, scene, and size. It is evident that ECF 2.0 is currently the largest available emotion cause dataset. 

Table \ref{tab:ECF2} presents the detailed statistics of our dataset for the two subtasks.
It can be seen that, in the entire ECF 2.0 dataset, 56.88\% of the utterances are labeled with one of the six basic emotions, 92.81\% of the emotion utterances have corresponding cause utterances, and 88.18\% of the emotion utterances are annotated with textual cause spans.

In addition, as shown in Figure \ref{fig:compare_conv_length} and Figure \ref{fig:compare_emotion}, the newly annotated test set is basically consistent with the original ECF 1.0 dataset in terms of conversation length and emotion distribution.

\begin{table*}[t]
\centering
\resizebox{\textwidth}{!}
{
    \begin{tabular}{cllccl}
    \hline
        \textbf{Rank} & \textbf{User Name} & \textbf{Team Name} & \textbf{w-avg. S. F$\rm _{1}$} & \textbf{w-avg. P. F$\rm _{1}$} & \textbf{Main Technologies} \\ \hline
        1 & Mercurialzs & Samsung Research China-Beijing\textsuperscript{$\dagger$} & 0.2300 & 0.3223 & LLaMA2, SpanBERT \\
        2 & sachertort & petkaz\textsuperscript{$\dagger$} & 0.1035 & 0.2640 & GPT 3.5, BERT \\ 
        3 & sharadC & UIC NLP GRADS\textsuperscript{$\dagger$} & 0.1839 & 0.2442 & RoBERTa, SpanBERT \\ 
        4 & nicolay-r & nicolay-r\textsuperscript{$\dagger$} & 0.1279 & 0.2432 & Flan-T5 \\
        5 & Mahshid & AIMA\textsuperscript{$\dagger$} & 0.0218 & 0.2102 & EmoBERTa, DeBERTa \\
        6 & jimar & UWBA\textsuperscript{$\dagger$} & 0.0639 & 0.2084 & RoBERTa, BERT \\
        7 & Choloe\_guo & UIR-ISC\textsuperscript{$\dagger$} & 0.1518 & 0.1963 & BERT, SpanBERT \\
        8 & aranjan25 & -- & 0.1431 & 0.1930 & -- \\
        9 & anaezquerro & LyS\textsuperscript{$\dagger$} & 0.0677 & 0.1823 & BERT \\
        10 & wrafal & PWEITINLP\textsuperscript{$\dagger$} & 0.0449 & 0.0723 & GPT-3, SpanBERT  \\
        11 & ericcui & \raisebox{-0.5ex}{\includegraphics[height=1em]{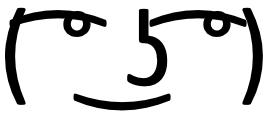}}-GPT & 0.0033 & 0.0339 & -- \\
        12 & conner & -- & 0.0000 & 0.0063 & -- \\
        13 & hpiotr6 & -- & 0.0000 & 0.0046 & -- \\
        14 & deliagrigorita & -- & 0.0005 & 0.0024 & -- \\ 
        15 & jpcf12 & VerbaNexAI Lab\textsuperscript{$\dagger$} & 0.0000 & 0.0000 & Logistic Regression, SpaCy \\ 
        \hline
\end{tabular}
}
\caption{The leaderboard for Subtask 1 (TECPE). ``$\dagger$'' indicates that the team has submitted a system description paper to SemEval-2024.}
\label{tab:subtask1_leaderboard}
\end{table*}

\begin{table*}[t]
\centering
\resizebox{\textwidth}{!}
{
    \begin{tabular}{cllccl}
    \hline
        \textbf{Rank} & \textbf{User Name} & \textbf{Team Name} & \textbf{w-avg. F$\rm _{1}$} & \textbf{Modality} & \textbf{Main Technologies} \\ \hline
        1 & Mercurialzs & Samsung Research China-Beijing\textsuperscript{$\dagger$} & 0.3774 & T,A,V & LLaMA2, RoBERTa, LLaVA \\
        2 & ZhanG\_XD & NUS-Emo\textsuperscript{$\dagger$} & 0.3460 & T,V & ChatGLM3 \\
        3 & SZTU-MIPS & SZTU-MIPS\textsuperscript{$\dagger$} & 0.3435 & T,A,V & MiniGPT-v2 \\
        4 & arefa & JMI\textsuperscript{$\dagger$} & 0.2758 & T,V & GPT-4V, GPT-3.5 \\
        5 & Mahshid & AIMA\textsuperscript{$\dagger$} & 0.2584 & T & EmoBERTa \\
        6 & jimar & UWBA\textsuperscript{$\dagger$} & 0.2506 & T,A,V & RoBERTa, BERT \\
        7 & julia-bel & DeepPavlov\textsuperscript{$\dagger$} & 0.2057 & T,A,V & Video-LLaMA \\
        8 & akshettrj & LastResort\textsuperscript{$\dagger$} & 0.1836 & T & BiLSTM, CRF  \\
        9 & oliver\_wang & QFNU\_CS\textsuperscript{$\dagger$} & 0.1786 & T,A,V & BERT \\
        10 & MSurfer20 & -- & 0.1708 & -- & -- \\
        11 & ayushg2000 & -- & 0.1635 & -- & -- \\
        12 & Hidetsune & Hidetsune\textsuperscript{$\dagger$} & 0.1288 & T & SpaCy, BERT \\
        13 & DuyguA & D-NLP & 0.0521 & -- & -- \\
        14 & bbgame605065444 & NCL\textsuperscript{$\dagger$} & 0.0146 & T,A,V & MLP \\
        15 & joshuashunk & -- & 0.0008 & -- & -- \\ \hline
\end{tabular}
}
\caption{The leaderboard for Subtask 2 (MECPE). ``$\dagger$'' indicates that the team has submitted a system description paper to SemEval-2024.}
\label{tab:subtask2_leaderboard}
\end{table*}

\section{Evaluation}
\label{sec:evaluation}

Our SemEval task runs on CodaLab\footnote{\url{https://codalab.lisn.upsaclay.fr/competitions/16141}}.
We released the training data in September 2023, and notified participants to commence model development.
The evaluation phase began on January 16, 2024, and ended on January 31, 2024.
We mixed the extended test set (consisting of 341 conversations with emotion and cause annotations; the labels are not publicly available) with some noise data (containing 324 conversations, not intended for evaluation) and released them together.
Each team is allowed to submit their results up to three times a day.

\subsection{Evaluation Metrics}

We evaluate the emotion-cause pairs of each emotion category with F$\rm _{1}$ scores separately and further calculate a weighted average of F$\rm _{1}$ scores across the six emotion categories, denoted as ``\textbf{w-avg. F$\rm _{1}$}''. Specifically, for Subtask 1, which involves the textual cause span, we adopt two strategies to determine whether the span is extracted correctly: 
\begin{itemize}
\setlength\itemsep{-0.2em}
\item \textit{Strict Match}: A predicted span is regarded as correct if it's the same as one of the annotated spans;
\item \textit{Proportional Match}: Calculate the overlap proportion of the predicted span and the annotated one.
\end{itemize}
The evaluation metrics for the two strategies are ``w-avg. S. F$\rm _{1}$'' and ``w-avg. P. F$\rm _{1}$'', respectively. Taking into account the complexity of Subtask 1, we choose ``\textbf{w-avg. P. F$\rm _{1}$}'' as the main metric\footnote{Specific calculation details can be found on \href{https://github.com/NUSTM/SemEval-2024_ECAC/tree/main/CodaLab/evaluation}{GitHub}.} for the ranking.

\subsection{Baselines}
As mentioned in our previous work \cite{wang2022multimodal}, for Subtask 2 we also employed the BiLSTM-based ECPE-2steps model as our baseline system. 
Specifically, we maintain the validation set of the ECF 1.0 datset unchanged and merge the test set into the training set to train the model. The evaluation of the model predictions on the extended test set achieves a weighted average F$\rm _{1}$ of \textbf{0.1926}.

For Subtask 1, based on the same model, we just convert the cause extraction module in Step 1 from the cause utterance prediction to the prediction of the start index and end index within the utterance, then simply match the indexes as candidate cause spans, followed by emotion-cause pairing and filtering in Step 2. The evaluation result for the weighted average proportional F$\rm _{1}$ on the extended test set is \textbf{0.1801}.

\subsection{Participating Systems and Results}
Our competition was created on Codalab in November 2023, and has attracted 143 registrations and a total of 216 submissions. 
After the evaluation, 18 teams have submitted system description papers. 

Team \textit{Samsung Research China-Beijing} \cite{Samsung2024SemEval} won first place in both subtasks, holding a significant lead over the second-place team. Teams \textit{petkaz} \cite{PetKaz2024SemEval} and \textit{UIC NLP GRADS} \cite{UIC2024SemEval} respectively captured the second and third places in Subtask 1. Teams \textit{NUS-Emo} \cite{NUS2024SemEval} and \textit{SZTU-MIPS} \cite{MIPS2024semeval} attained second and third positions in Subtask 2. The official leaderboards for Subtask 1 and Subtask 2 are shown in Table \ref{tab:subtask1_leaderboard} and Table \ref{tab:subtask2_leaderboard}, respectively.

\subsubsection{System Overview}
Almost all systems have implemented our task through a two-step framework, first performing the ERC task and then predicting the causes based on emotions.
In the following, we briefly introduce the systems from the top teams and some other notable approaches.

Team \textit{Samsung Research China-Beijing} \cite{Samsung2024SemEval} achieved first place in both subtasks with a pipeline framework. They fine-tuned the LLaMA2-based InstructERC \cite{lei2023instructerc} to extract the emotion category of each utterance in a conversation. For further data augmentation, they added three additional auxiliary tasks based on the original training data strategy of InstructERC. 
Then, the MuTEC \cite{bhat2023multi} and TSAM \cite{zhang2022tsam} models are used, respectively, to extract cause spans for Subtask 1 and cause utterances for Subtask 2.
They also obtained different multimodal representations through openSMILE \cite{eyben2010opensmile}, LLaVA \cite{liu2024llava}, and a self-designed face module to explore the integration of audio-visual information.
It should be noted that they used various models for ensemble learning to determine the final prediction.

Team \textit{petkaz} \cite{PetKaz2024SemEval} ranked second in Subtask 1. They fine-tuned GPT 3.5 \cite{Ouyang2022} for emotion classification and then used a BiLSTM-based neural network to detect cause utterances. The cause extractor model is initialized with BERT \cite{Devlin2019BERTPO}, followed by three BiLSTM layers. They treat the entire cause utterance as a cause span.

Team \textit{NUS-Emo} \cite{NUS2024SemEval} achieved the second highest score in Subtask 2. First, they conducted zero-shot testing experiments to evaluate multiple LLMs, including OPT-IML3 \cite{Iyer2022}, Instruct-GPT4 \cite{Peng2023}, Flan-T5 \cite{Chung2022}, and ChatGLM \cite{du2022glm}.  ChatGLM3-6B is ultimately selected as its backbone model based on its superior performance. They designed an emotion-cause-aware instruction-tuning mechanism to update the LLM and incorporated the visual representation from the ImageBind \cite{girdhar2023imagebind} encoder.

Team \textit{UIC NLP GRADS} \cite{UIC2024SemEval} achieved the third place in Subtask 1, and their system performed well in the strict metric, ranking second. They fine-tuned RoBERTa \cite{Liu2019RoBERTaAR} for emotion classification, and then further fine-tuned a SpanBERT \cite{Joshi2019SpanBERTIP} model that had been fine-tuned in SQuAD 2.0 \cite{Rajpurkar2018KnowWY}, to predict cause spans in QA format.

Team \textit{SZTU-MIPS} \cite{MIPS2024semeval} ranked third in Subtask 2. They integrated text, audio, and image modalities for emotion recognition and adopted the MiniGPTv2 model \cite{Chen2023MiniGPTv2LL} for multimodal cause extraction.
Specifically, textual features are obtained from InstructERC, while acoustic features are extracted using HuBERT \cite{Hsu2021HuBERTSS}. For visual modality, faces are first extracted using OpenFace \cite{Baltruaitis2016OpenFaceAO} from video frames, followed by extraction of facial features using expMAE \cite{Cheng2023SemiSupervisedME}.

Team \textit{nicolay-r} \cite{Nicolay2024SemEval} finetuned Flan-T5 by designing the chain of thoughts for emotion causes based on the Three-Hop Reasoning (THOR) framework \cite{Fei2023ReasoningIS}, to predict the emotion of the current utterance and the emotion caused by the current utterance towards the target utterance. Their reasoning revision methodology and rule-based span correction technique bring further improvements.

Team \textit{JMI} \cite{JMI2024SemEval} implemented two different approaches. In their best system, they used in-context learning using GPT 3.5 for emotion prediction and cause prediction, respectively. 
Conversation-level video descriptions were extracted via GPT-4V \cite{yang2023dawn} to provide more context to GPT 3.5. In addition, they also fine-tuned two separate Llama2 \cite{Touvron2023Llama2O} models to recognize emotions and extract causes.

Team \textit{AIMA} \cite{AIMA2024SemEval} fine-tuned EmoBERTa \cite{Kim2021EmoBERTaSE} for emotion classification and then obtained the emotion-cause pairs via a Transformer-based encoder. After finding the pairs, they further fine-tuned the DeBERTa \cite{he2021debertav3} that had been fine-tuned on SQuAD 2.0 to extract the cause spans for Subtask 1.

Team \textit{UWBA} \cite{UWBA2024SemEval} fused the features of three modalities at the utterance level and then used them for emotion classification and pair prediction. It is interesting that they summarized five span categories (\textit{Whole Utterance}, \textit{First part}, \textit{Last part}, \textit{Middle part}, \textit{Other}) through observations of training data, and then trained a classifier to further predict textual cause spans in cause utterance.

Furthermore, Team \textit{DeepPavlov} \cite{DeepPavlov2024SemEval} investigated the performance of Video-LLaMA \cite{Zhang2023VideoLLaMAAI} in several modes and found that model fine-tuning yields notable improvements in emotion and cause classification. 
Team \textit{PWEITINLP} \cite{PWEITINLP2024SemEval} utilized GPT-3 for emotion classification. Some other Teams, including \textit{UIR-ISC} \cite{UIR2024SemEval}, \textit{LyS} \cite{LyS2024SemEval}, \textit{QFNU\_CS} \cite{QFNU2024SemEval} and \textit{Hidetsune} \cite{Hidetsune2024SemEval}, all employed BERT-based models to address our task, among which \textit{LyS} proposed an end-to-end model comprising a BERT encoder and a graph-based decoder to identify emotion cause relations.
Team \textit{LastResort} \cite{LastResort2024SemEval} tackled our task as sequence labeling problems and used BiLSTM followed by a CRF layer to solve it.
Team \textit{NCL} \cite{NCL2024SemEval} solely utilized pre-trained models to extract features from three modalities.
Team \textit{VerbaNexAI Lab} \cite{VerbaNexAI2024SemEval} demonstrated the inadequacy of machine learning techniques alone for emotion cause analysis.

\subsubsection{Discussion}

Our task, Multimodal Emotion Cause Analysis in Conversations, involves informal real-life conversations and complex audio-visual scenes. Additionally, emotions exhibit strong subjectivity, and we have observed that even humans sometimes struggle to accurately identify emotions and their causes. This complexity underscores the intricate nature of human emotions and the nuanced contexts in which they occur, posing a substantial challenge for data annotation and subsequent model development.

\paragraph{Dataset Bias.}
Emotion category imbalance is an inherent problem in the ERC task \cite{li2017dailydialog,hsu2018emotionlines,poria2019meld}, aligning with real-world phenomena where people tend to express positive emotions like \textit{joy} more frequently in their daily communications, while expressions of \textit{disgust} and \textit{fear} are less common. 
Our dataset is sourced from TV series that closely resemble the real world, naturally also exhibiting an imbalance in emotions, as illustrated in Figure \ref{fig:compare_emotion}.
However, such an imbalance may adversely affect a model's ability to learn and generalize across different emotions, potentially leading to biases towards frequently expressed emotions \cite{PetKaz2024SemEval,UIC2024SemEval}. 
Moreover, emotion cause datasets often have a noticeable pattern in the location of causes and emotions. Some systems rely on this position bias, either by using a fixed window size or by direct post-processing to add the emotion utterance as the cause \cite{Nicolay2024SemEval,JMI2024SemEval}, which overlooks the effective semantic connections between distant contexts and may lead to poor generalization capabilities for unseen data where the cause is not in proximity to the emotion.
In the future, LLMs can be leveraged to assist with annotation to expand the diversity of datasets available for fine-tuning, which encompass a wider range of emotional expressions and cultural backgrounds. This can mitigate existing dataset biases and enhance the model's applicability and generalizability across various scenarios.

\paragraph{Utilization of LLMs.}
Recently, LLMs have exhibited remarkable capabilities in a wide range of tasks and are rapidly advancing the field of natural language processing. Therefore, LLMs are allowed to be used in our competition.
It is evident that about a third of the teams have used LLMs for emotion cause analysis, and most of them are ranked at the top.
However, some participants have observed that LLMs perform poorly in zero-shot and few-shot settings on emotion and cause recognition tasks \cite{PetKaz2024SemEval,JMI2024SemEval,DeepPavlov2024SemEval}, indicating a crucial need for task-specific fine-tuning. Furthermore, prompt engineering is essential, as LLMs often produce hallucinations or unstructured outputs. Due to resource and cost constraints, most researchers cannot take full advantage of the strongest capabilities of LLM. Future research is encouraged to explore ways to enhance lightweight models or to bridge the gap between pre-training and downstream tasks, thereby augmenting LLMs' ability to understand emotions.

\paragraph{Potential of Multimodal Information.}
Multimodal information is important for discovering both emotions and their causes in conversations. In our daily communications, we depend not only on the speaker’s voice intonation and facial expressions to perceive his emotions, but also on some auditory and visual scenes to speculate the potential causes that trigger the emotions of speakers beyond text. 
However, some participants found that the introduction of audio or visual modalities results in minimal improvements or even a decrease in system performance \cite{Samsung2024SemEval,MIPS2024semeval,UWBA2024SemEval}. 
This issue arises partly due to the characteristics of our dataset, which involves a large number of complex visual scenes but few visual cause clues, leading to the introduction of noise. 
Another limiting factor might be that multimodal feature extraction methods are not advanced enough or fusion strategies are not effective enough.
The challenges that require further exploration include the effective interaction and fusion of multimodal information, as well as the perception, understanding, and utilization of audiovisual scenes. 
Furthermore, there is a demand for more high-quality data sets on multimodal emotion cause analysis to support research in this area.

\section{Conclusions}
In this paper, we describe the SemEval-2024 Task 3 named Multimodal Emotion Cause Analysis in Conversations, which aims to extract all potential pairs of emotions and their corresponding causes from a conversation. 
The shared task has attracted 143 registrations and 216 successful submissions.
We provide detailed descriptions of task definition and data annotation, summarize participating systems, and discuss their findings.

As an important direction of affective computing, multimodal emotion cause analysis in conversation plays an important role in many real-world applications. We hope that our research and resources can contribute towards the design of future systems in this direction.

\section{Ethics Statement} 
Our \textbf{ECF 2.0} dataset is annotated on the basis of the MELD dataset \footnote{\url{https://github.com/declare-lab/MELD}} which is licensed under the GNU General Public License v3.0 and is used only for scientific research.
We do not share personal information and do not release sensitive content that can be harmful to any individual or community.
Conducting multimodal emotion cause analysis will help us better understand emotions in human conversations, build human-machine dialogue systems, and contribute to society and human well-being.

\section*{Acknowledgements}

We express our sincere gratitude to the annotators who contributed to constructing the dataset, laying a solid foundation for our research. 
We also extend our heartfelt gratitude to all the participants who participated in our competition, especially the teams who submitted system description papers and completed the reviewing tasks assigned to them. 
Furthermore, we thank the anonymous reviewers for their invaluable feedback and insightful comments.

\bibliography{custom}

\newpage

\appendix


\end{document}